\documentclass{article}

\usepackage{arxiv}

\usepackage[utf8]{inputenc} 
\usepackage[T1]{fontenc}    
\usepackage{hyperref}       
\usepackage{url}            
\usepackage{booktabs}       
\usepackage{amsfonts}       
\usepackage{amsmath,amssymb,amsthm}        
\usepackage{nicefrac}       
\usepackage{microtype}      
\usepackage{graphicx}
\usepackage{lipsum}
\usepackage{lmodern}
\usepackage{geometry}
\usepackage{color}
\usepackage{subfigure}
\usepackage{enumerate}
\usepackage[capitalize]{cleveref}
\usepackage[ruled,vlined]{algorithm2e}

\newcommand{\bvec}[1]{\mathbf{#1}}

\newcommand{\mc}[1]{\mathcal{#1}}

\newcommand{\vk}{\bvec{k}}

\newcommand{\vw}{\bvec{w}}
\newcommand{\vx}{\bvec{x}}
\newcommand{\ud}{\,\mathrm{d}}

\newcommand{\norm}[1]{\left\lVert#1\right\rVert}
\newcommand{\wt}[1]{\widetilde{#1}}

\newcommand{\Or}{\mathcal{O}}

\newcommand{\RR}{\mathbb{R}}

\DeclareMathOperator*{\argmin}{arg\,min}

\newtheorem{thm}{\protect\theoremname}
\theoremstyle{plain}

\theoremstyle{plain}

\theoremstyle{plain}
\newtheorem*{lem*}{\protect\lemmaname}
\theoremstyle{plain}
\newtheorem{prop}[thm]{\protect\propositionname}
\theoremstyle{plain}

\newtheorem{defn}[thm]{Definition}

\providecommand{\corollaryname}{Corollary}
\providecommand{\lemmaname}{Lemma}
\providecommand{\propositionname}{Proposition}
\providecommand{\remarkname}{Remark}
\providecommand{\theoremname}{Theorem}

\newcommand{\Rev}[1]{\textcolor{black}{#1}}

\title{Learning the mapping \(\mathbf{x}\mapsto \sum_{i=1}^d x_i^2\):
the cost of finding the needle in a haystack}

\author{
Jiefu Zhang\textsuperscript{*}\\
Department of Mathematics,\\ 
University of California, Berkeley,\\ 
Berkeley, CA 94720.\\ 
\texttt{jiefuzhang@berkeley.edu}
\And
Leonardo Zepeda-N\'u\~nez\\
Department of Mathematics, \\
University of Wisconsin-Madison,\\
Madison, WI 53706.\\ 
\texttt{lzepeda@math.wisc.edu}
\AND
Yuan Yao\\
Department of Mathematics, \\
Hong Kong University of Science and Technology,\\
Clear Water Bay, Kowloon, Hong Kong SAR. \\ 
\texttt{yuany@ust.hk}
\AND
Lin Lin\\
Department of Mathematics, \\
University of California, Berkeley, \\
Computational Research Division, \\
Lawrence Berkeley National Laboratory, \\
Berkeley, CA 94720. 
\\ \texttt{linlin@math.berkeley.edu}
}

\begin{document}

\maketitle 
\begin{abstract}
The task of using machine learning to approximate the mapping $\vx\mapsto\sum_{i=1}^d x_i^2$ with $x_i\in[-1,1]$ seems to be a trivial one. Given the knowledge of the separable structure of the function, one can design a sparse network to represent the function very accurately, or even exactly. When such structural information is not available, and we may only use a dense neural network, the optimization procedure to find the sparse network embedded in the dense network is similar to finding the needle in a haystack, using a given number of samples of the function.
We demonstrate that the cost (measured by sample complexity) of finding the needle is directly related to the Barron norm of the function. While only a small number of samples is needed to train a sparse network, the dense network trained with the same number of samples exhibits large test loss and a large generalization gap. In order to control the size of the generalization gap, we find that the use of explicit regularization becomes increasingly more important  as $d$ increases. The numerically observed sample complexity with explicit regularization scales as $\mathcal{O}(d^{2.5})$, which is in fact better than the theoretically predicted sample complexity that scales as $\mathcal{O}(d^{4})$.  Without explicit regularization (also called implicit regularization), the numerically observed sample complexity is significantly higher and is close to  $\mathcal{O}(d^{4.5})$.
\end{abstract}

\section{Introduction}\label{sec:intro}

Machine learning and, in particular, deep learning methods have revolutionized numerous fields such as speech recognition \cite{Hinton2012}, computer vision \cite{Krizhevsky2012}, 
drug discovery \cite{MaSheridan2015}, genomics \cite{Leung2014}, etc. The foundation of deep learning is the universal approximation theorem~\cite{CohenSharir2018,Hornik91,Khrulkov2018,Mhaskar2018}, which allows neural networks (NN) to approximate a large class of functions arbitrarily well, given a sufficient large number of degrees of freedom. In practice, however, the number of degrees of freedom is often limited by the computational power, thus the choice of the architecture to reduce the number of degrees of freedom is of paramount importance for the quality of the approximation~\cite{He2015ConvolutionalNN,He2016DeepRL,Mhaskar2018}. Empirically, NN models have been shown to be surprisingly efficient in finding good local, and sometimes global, optima when using an overparameterized model, e.g. training a sparse teacher network is less efficient than training a dense, overparametrized student network \cite{Livni2014}. It has been argued that the energy landscape of an overparameterized model may be benign, and in certain situations all local minima become indeed global minima~\cite{Bruna2017landscape,Kawaguchi2016,Ge2019landscape,Bruna2018landscape}. Furthermore,  starting from an overparameterized model, observations such as the lottery ticket hypothesis \cite{Frankle2019lotterya,Frankle2019lotteryb,Liu19pruning} states that with proper initializations, it is possible to identify the ``winning tickets'', i.e. a sparse subnetwork with accuracy comparable to the original dense network.  

We point out that many of the aforementioned studies focus on image classification problems using common data sets such as MNIST, CIFAR10 and ImageNet, with or without the presence of noise. However, in scientific computing, the setup of the problem can be very different: usually, we are interested in using NN models to parameterize a smooth, high-dimensional function accurately, and often without artificial noise. Within this context, the results mentioned above naturally raise the following questions: 
\begin{enumerate}[ {(}1{)} ]
\item  How important is it to select the optimal architecture? In other words, does it matter whether one uses an overparameterized model? 

\item  If there is a sparse subnetwork that is as accurate as the overparameterized network, can the training procedure automatically identify the subnetwork? In other words, what is the cost of finding the needle (sparse subnetwork) in a haystack (overparameterized network)? 

\item If (2) is possible, how does the training procedure (such as the use of regularization) play a role?
\end{enumerate}

This paper presents a case study of these questions in terms of a deceivingly simple task: given \(\vx\in [-1,1]^d\) drawn from a certain probability distribution and a target accuracy $\epsilon$, learn the square of its 2-norm, i.e. the function
\begin{equation}
\wt{f}^*(\vx):=\sum_{i=1}^d x_i^2.
\label{eqn:fx}
\end{equation}
More specifically, for a given neural network model $f(\vx,\theta)$, where $\theta$ denotes the parameters in the model, and for a given loss function, such as the quadratic loss
\(\ell\left(y, y^{\prime}\right)=\frac{1}{2}\left(y-y^{\prime}\right)^{2}\), our goal is to find $\theta$ such that the population loss
\[
L(\theta)=\mathbb{E}_{\Rev{\vx}}[\ell(f(\vx; \theta), \wt{f}^{*}(\vx))]\le \epsilon.
\]
Note that  $\wt{f}^*(\vx)\sim \Or(d)$, so we consider the scaled target function \footnote{Correspondingly $\wt{f}^*$ will be called the original target function, or the unscaled target function.} in order to normalize the output 
\begin{equation}
f^*(\vx)=\frac{1}{d}\wt{f}^*(\vx).
\label{eqn:fx_scaled}
\end{equation}
However, as in many scientific computing applications, the magnitude of the quantity of interest indeed grows with respect to the dimension, and our interest here is to approximate the original function $\wt{f}^*(\vx)$ to $\epsilon$ accuracy. Using a quadratic loss the population loss needed for approximating $f^*$ becomes $\epsilon/d^2$. In other words, if each component of $\vx$ is chosen randomly, then by the law of large number $f^*(\vx)$ converges to a constant $\mathbb{E}[ x^2]$ as $d\to \infty$. So the  $\epsilon/d^2$ target accuracy means that it is the deviation from such a mean value that we are interested in.  

If we are allowed to use the mapping \(x\mapsto x^2\) as an activation function, we would apply this function to each component and sum up the results, ensuing that the representation will be exact. Therefore, we exclude such an activation function, and only use standard activation functions such as ReLU or sigmoid, which requires only $\Or(\log (1/\epsilon))$ neurons to reach accuracy  $\epsilon$~\cite{Yarotsky2017}. 

\begin{figure}[ht] 
\centering 
\includegraphics[trim=0mm 0mm 0mm 40mm,clip,width=0.45\textwidth]{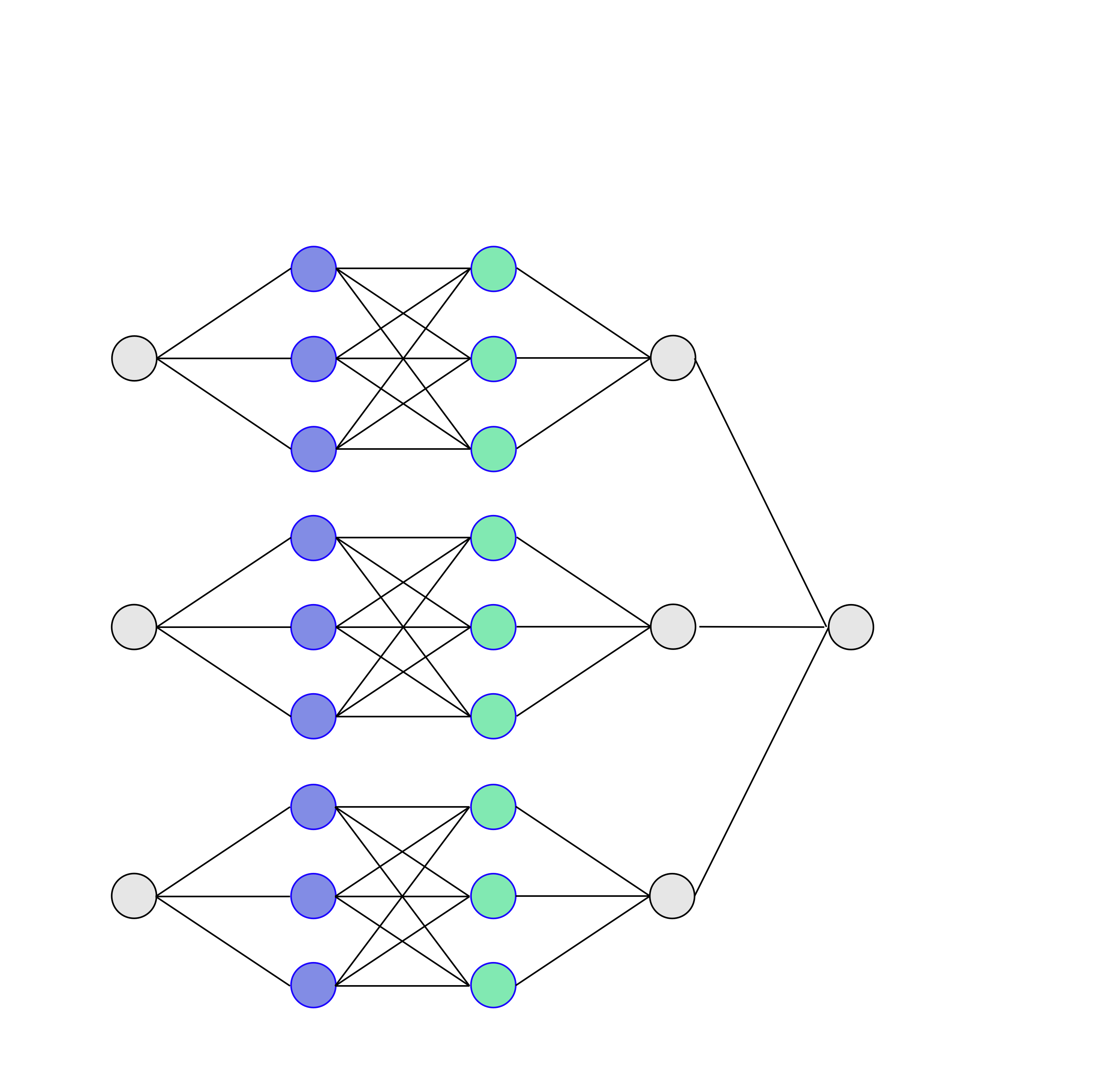}
\includegraphics[trim=0mm 0mm 0mm 40mm,clip,width=0.45\textwidth]{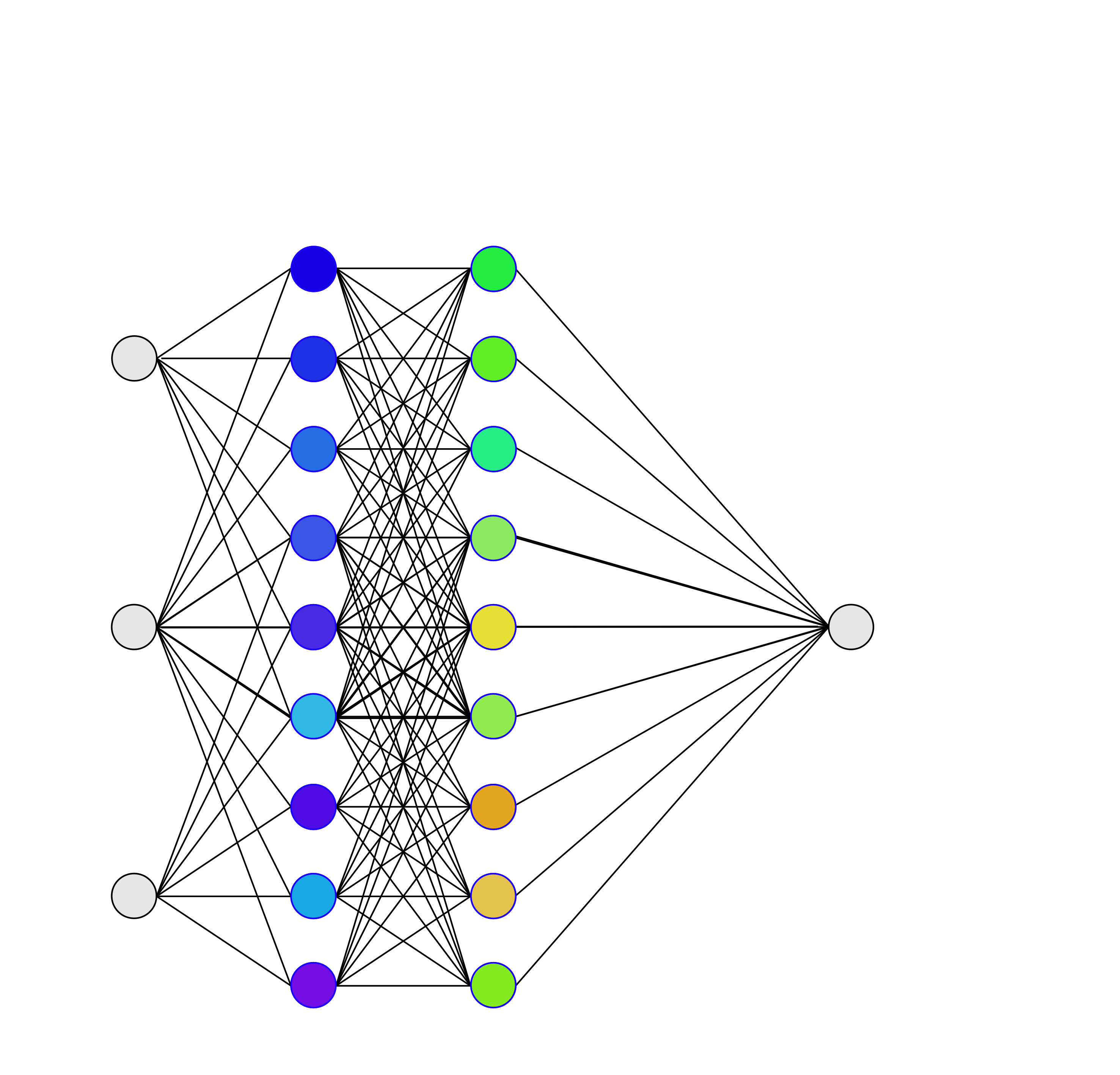}
\caption{The architecture of the local network is on the left, and the architecture of the global network is on the right. Here $d=3$ and the number of channels $\alpha = 3$.} 
\label{fig:architecture} 
\end{figure}

We can build a network leveraging the separability of Eq.~\eqref{eqn:fx}. In particular, we can use a small network to approximately represent the scalar mapping \(x\mapsto x^2\), and sum up the results from all components. The weights for the neural network of each component are shared, so the number of parameters is independent of $d$. The network will be called the local network (LN) below. \Rev{However,} if we do not have the \textit{a priori} structural information that the target function is separable, we need to use a dense or fully connected neural network, which is referred to as the global network (GN). \cref{fig:architecture} sketches the structure of LN and GN. Note that LN can be naturally embedded into GN as a subnetwork by deleting certain edges. Therefore the \textit{optimal} performance of GN should be at least as good as that of LN.

Throughout the paper we assume that the number of neurons in the LN is large enough so that the scalar mapping $x\mapsto x^2$ is learned very accurately (with test error less than e.g. $10^{-5}$), and the structure GN is then obtained by connecting all the remaining edges. In addition to the architectural bias, we are also concerned with the sample complexity about GN, i.e. the number of independent samples of $f(\vx)$ as the training data to reach a certain target accuracy (i.e. generalization error). Our results can be summarized as follows.

\begin{enumerate}

\item Using the same number of samples, LN can perform significantly better than GN. This shows that the global energy landscape of GN cannot be very simple, and the desired LN subnetwork cannot be easily identified.

\item If we embed a converged LN into a GN, add small perturbations to the weights of GN, and start the training procedure, LN can still outperform GN. This shows that the local energy landscape of GN may not be simple either.

\item In order to use GN to achieve performance that is comparable to LN, we need a significantly larger number of samples. The number of samples needed to reach certain target accuracy increases with respect to the dimension as $\Or(d^\gamma)$ up to logarithmic factors. From \textit{a priori} error analysis, we have $\gamma=4$. 

\item The numerical scaling of the sample complexity with respect to $d$ depends on the regularizations.
In particular, when proper regularization ($\ell^1, \ell^2$, or path norm  regularization \cite{EMaWu2018}) is used, the numerically observes sample complexity is around $\Or(d^{2.5})$, which behaves better than the  theoretically predicted worst case complexity, which scales as $\Or(d^4)$ . On the other hand, when implicit regularization (i.e. early stopping \cite{onEarlyStopping}) is used, we observe $n\sim \Or(d^{4.5})$ , i.e. the sample complexity using implicit regularization is significantly larger than that with explicit regularization. The early stopping criteria we use in this paper is that: after $T$ (1000 in default) epochs, we find an optimal $t^\ast\leq T$ where the validation error is minimized. 
Furthermore, the trained weight matrix obtained with explicit regularization is approximately a sparse matrix, while the weight matrix obtained with early stopping is observed to be a dense matrix.

\end{enumerate}

\noindent\textbf{Related works:}

In order to properly describe the sample complexity to reach certain target accuracy, we need to have \textit{a priori} error estimate (a.k.a. worst-case error), and/or \textit{a posteriori} error estimate (a.k.a. instance-based error) of the generalization error. For two-layer neural networks, such estimates have been recently established \cite{Bach2017,EMaWu2018} for a large class of functions called Barron functions \cite{Barron1993,KlusowskiBarron2016}. The estimates have also been recently extended to deep networks based on the ResNet structure \cite{He2016DeepRL,EMaWang2019}. We obtain our theoretical estimate of the sample complexity with respect to $d$ by applying results of \cite{EMaWu2018} to  $f(\vx)$. For a sufficiently wide two-layer neural network, \cite{AroraDuHuEtAl2019} studied the generalization error together with the gradient descent dynamics when using a polynomial activation function. The problem of ``finding the needle in a haystack'' by comparing the performance of fully-connected networks (FCN) and convolutional neural networks (CNN) was also recently considered by \cite{AscoliSagunBrunaEtAl2019} for image recognition problems. It was found that there are rare basins in the space of fully-connected networks associated with very small generalization errors, which can be accessed only with prior information from CNN. This corroborates our finding for learning $f(\vx)$ here. Note that the separable structure in the target function can also be viewed from the perspective of permutation symmetry. Therefore our study also supports the argument for recognizing the importance of preserving symmetries when designing neural network architectures, which has been observed by numerous examples in physics based machine learning applications \cite{ZaheerKotturRavanbakhshEtAl2017,ZhangHanWangCarE2018}.  
Our result also corroborates the recent study \cite{NagarajanKolter2019}, which questioned the prediction power of theoretical generalization error bound rates for overparameterized deep networks trained without explicit regularization. 

\subsection{Organization}
In Section \ref{sec:theory} we provide the theoretical foundations for the two-layer networks, including {\it a priori} and {\it a posteriori} bounds on the generalization gap. In Section \ref{sec:LN_GN} we use the theory developed in Section \ref{sec:theory} to find the bounds for the generalization error for the squared norm. Section \ref{sec:numer} provides the numerical experiments, followed by discussion in Section \ref{sec:conclusion}. 

\section{Generalization error of two-layer networks}\label{sec:theory}

In this section, we briefly describe the concept of the Barron norm, and the generalization error for two-layer neural networks. We refer readers to \cite{EMaWu2018,EMaWu2019} for more details. Let the domain of interest be \(\Omega=[-1,1]^d\). We assume the magnitude of the target function is already normalized to be $\Or(1)$, e.g. the scaled function $f^*(\vx)$ in \cref{eqn:fx_scaled}. Then for any $y'\sim \Or(1)$, both the magnitude and the Lipschitz constant of the square loss function $\ell(\cdot,y')$ are of $\Or(1)$. 

\subsection{Barron norm}

We say that a function
\(f:\Omega\to\RR\) can be represented by a two-layer NN if
\begin{equation}
f(\vx; \theta)=\sum_{k=1}^{m} a_{k} \sigma\left(\vw_{k}^{\top} \vx\right).
\label{eqn:two_layer}
\end{equation}
Here \(\vw_k\in\RR^d\), \(\theta:=\{(a_k,\vw_k)\}_{k=0}^m\) represents all the
parameters in the network, and \(\sigma(\cdot)\) is an scale-invariant
activation function such as ReLU. The scale invariance implies
\begin{equation}
\sigma(\vw^{\top}\vx)=\norm{\vw}_1\sigma(\hat{\vw}^{\top}\vx), \quad \hat{\vw}=\vw/\norm{\vw}_1.
\label{eqn:}
\end{equation}
Therefore we may, without loss of generality, absorb the magnitude $\norm{\vw}_1$ into the scalar $a$, and assume $\norm{\vw}_1=1$.  

The training set is composed of \(n\)
i.i.d. samples \(\mc{S}=\{(\vx^{(i)},y^{(i)})\}_{i=1}^n\). To distinguish the indices, we use \(\vx^{(i)}\) to denote the \(i\)-th sample of the vector
\((1\le i\le n)\), and use \(x^{(i)}_j\) to denote the \(j\)-th component of
the vector \(\vx^{(i)}\) \((1\le j\le d)\).

Our goal is to minimize the population loss
\[L(\theta)=\mathbb{E}_{\vx}[\ell(f(\vx; \theta), y)],\]
through the minimization of the training loss

\[\hat{L}_{n}(\theta)=\frac{1}{n} \sum_{i=1}^{n}
\ell\left(f\left(\vx^{(i)} ; \theta\right), y^{(i)}\right).\]
For a realization of parameters $\theta$,  the generalization gap is defined as \(\left|L(\theta)-\hat{L}_{n}(\theta)\right|\).

A function \(f\) represented by a two-layer neural network is a special
case of the Barron function, which admits the following integral
representation
\begin{equation}
f(\vx)=\int_{S^{d}} a(\vw) \sigma(\langle \vw, \vx\rangle) \ud \pi(\vw),
\label{eqn:twolayer_cont}
\end{equation}
where \(\pi\) is a probability distribution over
\(S^d:=\{\vw|\norm{\vw}_1=1\}\), and \(a(\cdot)\) is a scalar function. In particular, when we choose $\pi(\vw):=\sum_{k=1}^m \delta(\vw-\vw_k)$ to be a discrete measure and define $a_k=a(\vw_k)$, we recover the standard two-layer network in \cref{eqn:two_layer}.

\begin{defn}[Barron norm and Barron space]  
 Let \(f\) be a Barron function. Denote
by \(\Theta_f\) all the possible representations of \(f\), i.e.
\[\Theta_f=\{(a,\pi)|f(\vx)=\int_{S^{d}} a(\vw) \sigma(\langle \vw, \vx\rangle) \ud \pi(\vw)\}.\]
Then the Barron-p norm is defined by
\begin{equation}
\gamma_p(f):=\inf _{(a, \pi) \in \Theta_{f}}\left(\int_{S^{d}}|a(\vw)|^{p} \ud \pi(\vw)\right)^{1 / p}.
\label{eqn:barron_norm}
\end{equation}
We may then define the Barron space as
\begin{equation}
\mathcal{B}_{p}(\Omega)=\left\{f : \gamma_p(f)<\infty\right\}.
\label{eqn:barron_space}
\end{equation}
\end{defn}

Unlike the standard $L^p$ norms, \cref{prop:barronspace} shows a remarkable result, which is that all Barron norms are equivalent (\cite[Proposition 2.1]{EMaWu2019}).
\begin{prop}[Equivalence of Barron norms]
\label{prop:barronspace}
For any function $f\in \mc{B}_1(\Omega)$, 
\begin{equation}
\gamma_1(f)=\gamma_p(f), \quad 1\le p\le \infty .
\label{eqn:gamma_equivalency}
\end{equation}
\end{prop}

To see why this can be the case, let us consider the two-layer NN in \cref{eqn:two_layer}, and assume $\norm{\vw_k}_{1}=1$ for all $k$. By H\"older's inequality  $\gamma_p(f)\le \gamma_{q}(f)$ when $1\le p<q\le \infty$. In particular, $\gamma_1(f)\le \gamma_\infty(f)$, and we only need to show $\gamma_\infty(f)\le \gamma_1(f)$.  Since $\pi$ should be a discrete measure, let $(a,\pi)$ be the minimizer for $\gamma_1$ as in \cref{eqn:barron_norm}. Then (recall $a_k:=a(\vw_k)$)
\begin{displaymath}
\gamma_1(f)= \sum_{k=1}^{m}\left|a_{k}\right|.
\end{displaymath}  
However, we may define a new distribution 
\begin{displaymath}
\wt{\pi}(\vw)= \sum_{k=1}^{m}\frac{1}{\gamma_1(f)}\left|a_{k}\right|\delta(\vw-\vw_k),
\end{displaymath}
then
\begin{displaymath}
\int_{S^d} a(\vw) \ud\pi(\vw)= \sum_{k=1}^{m}\left|a_{k}\right| =\gamma_1(f) \sum_{k=1}^m  \frac{1}{\gamma_1(f)}\left|a_{k}\right|=\gamma_1(f)\int_{S^d} \ud\wt{\pi}(\vw).
\end{displaymath}
In other words, $(\wt{a},\wt{\pi})\in \Theta_f$, where $\wt{a}_k=\gamma_1(f)$ is a constant. By definition of \cref{eqn:barron_norm},
\begin{displaymath}
\gamma_{\infty}(f)\le \gamma_1(f) \int_{S^d} \ud \wt{\pi}(\vw) = \gamma_1(f).
\end{displaymath}
This proves $\gamma_{\infty}(f)=\gamma_1(f)$. The same principle can be generalized to prove \cref{prop:barronspace} for general Barron functions \cite{EMaWu2019}. 

\cref{prop:barronspace} shows that the definition of the Barron space $\mc{B}_p(\Omega)=\mc{B}_1(\Omega)$ is independent of $p$. Therefore we may drop the subscript $p$ and write  $\mc{B}_p(\Omega)$ as $\mc{B}(\Omega)$. Similarly we denote by $\norm{f}_{\mc{B}}:=\gamma_1(f)$ as the Barron norm.

\subsection{Error bound}

For a given parameter set $\theta$ defining \cref{eqn:two_layer}, the Barron norm is closely related to the path norm 
\begin{equation}
\|\theta\|_{\mathcal{P}} := \sum_{k=1}^{m}\left|a_{k}\right|.
\label{eqn:path_norm}
\end{equation}

According to \cref{eqn:barron_norm} we immediately have $\norm{f}_{\mc{B}}\le \|\theta\|_{\mathcal{P}}$. The path norm can be used to obtain the following \textit{a posteriori} error estimate for the generalization gap for any choice of $\theta$.

\begin{thm}[\textit{A posteriori} error estimate \cite{EMaWu2018}] 
\label{thm:apost}
For any choice of
parameter \(\theta\), for any \(\delta>0\), and with probability at
least \(1-\delta\) over the choice of the training set \(\mc{S}\),
\[\left|L(\theta)-\hat{L}_{n}(\theta)\right| \lesssim   \sqrt{\frac{\ln (d)}{n}}\left(\|\theta\|_{\mathcal{P}}+1\right)
+ \sqrt{\frac{\ln ((\|\theta\|_{\mathcal{P}}+1)^{2} / \delta)}{n}}.\]

\end{thm}

For a given two-layer network, the path norm can be computed directly. Hence in order to reduce the generalization error to $\epsilon$, the number of samples needed is approximately $\Or(\|\theta\|_{\mathcal{P}}^{2}/\epsilon^{2})$, which is a practically useful bound. However, in order to estimate the scaling of $n$ with respect to the dimension $d$ for a particular function, we need to replace the $\|\theta\|_{\mathcal{P}}$ by some measure of the complexity associated with $f^*$ itself, such as the Barron norm. There indeed exists a two-layer network with a bounded path norm, so that the population loss is small. This is given in \cref{prop:bound_path} (\cite[Proposition 2.1]{EMaWu2018}).

\begin{prop}
\label{prop:bound_path}
For any $f \in \mathcal{B}(\Omega),$ there exists a two layer neural network $f(\vx; \tilde{\theta})$ of width $m$ with $\|\tilde{\theta}\|_{\mathcal{P}} \leq$
$2 \norm{f^*}_{\mc{B}}$, such that
\begin{equation}
L(\wt{\theta}) \lesssim  \frac{\norm{f^*}_{\mc{B}}}{m}.
\label{eqn:poploss_approx}
\end{equation}
\end{prop}

\cref{eqn:poploss_approx} characterizes the approximation error due to the use of a neural network of finite width, which decays as $m^{-1}$. \cref{prop:bound_path} states that it is in principle possible to reduce the population loss while keeping the path norm being bounded.  However, numerical results (both previous works and our own results here) indicate that when minimizing with respect to the training loss directly (when early stopping is used, this is also called the implicit regularization), the path norm associated with the optimizer can be very large. According to \cref{thm:apost}, the resulting generalization error bound can be  large as well.

A key result connecting the \textit{a priori} and \textit{a posteriori} error analysis is to impose stronger requirements of the training procedure. Instead of minimizing with respect to the training loss $\hat{L}_{n}(\theta)$ directly,
\cite{EMaWu2018} proposes to minimize with respect to the following regularized loss function
\begin{equation}
J_{\lambda}(\theta) :=\hat{L}_{n}(\theta)+\lambda\|\theta\|_{\mathcal{P}},
\label{eqn:regularize_loss}
\end{equation}
where \(\lambda>0\) is a penalty parameter. The corresponding minimizer is defined as

\[\hat{\theta}_{n, \lambda}=\argmin_{\theta} J_{\lambda}(\theta).
\]
The benefit of minimizing with respect to \cref{eqn:regularize_loss} is that the path norm is penalized explicitly in the objective function, which allows us to control  both the path norm and the generalization error.

\begin{thm}[\textit{A priori} error estimate \cite{EMaWu2018}] 
\label{thm:apriori}
Assume that the target function
\(f^*\in \mc{B}(\Omega)\), and
\(\lambda\ge \lambda_n=4 \sqrt{2 \ln (2 d) / n}\). Then for
any \(\delta>0\) and with probability at least \(1-\delta\) over the
choice of the training set \(\mc{S}\)
\begin{equation}
\label{eqn:poploss_apriori}
L(\hat{\theta}_{n, \lambda})  \lesssim \frac{\norm{f^*}_{\mc{B}}^2}{m}+\lambda (\norm{f^*}_{\mc{B}}+1) +\frac{1}{\sqrt{n}}\left(\norm{f^*}_{\mc{B}}+\sqrt{\ln (n / \delta)}\right).
\end{equation}
The path norm of the parameter satisfies
\begin{equation}
\left\|\hat{\theta}_{n, \lambda}\right\|_{\mathcal{P}} \lesssim \frac{\norm{f^*}_{\mc{B}}^2}{\lambda m}+\norm{f^*}_{\mc{B}} +\sqrt{\ln (1 / \delta)}.
\label{eqn:path_reg_bound}
\end{equation}
\end{thm}

For a fixed target function $f^*$, the contribution to the error in \cref{eqn:poploss_apriori} can be interpreted as follows: the first
term is the approximation error, determined by the width of the network.
The third is determined by the sample complexity which is proportional
to \(n^{-\frac12}\). The second term is present due to the need of balancing the loss and the path norm in the objective function \eqref{eqn:regularize_loss}. If $\lambda$ is too large, then the regularized loss function is too far away from the training loss, and the error bound becomes large. On the other hand, \cref{thm:apriori} requires $\lambda$ should be at least  $\sim n^{-\frac12}$. This can also be seen from \cref{eqn:path_reg_bound} that if $\lambda$ is too small, the path norm becomes unbounded. Therefore to balance the two factors we should choose  the
regularization parameter as \(\lambda\sim n^{-\frac12}\).

\section{Generalization error for squared norm} \label{sec:LN_GN}

In this section we apply the generalization error bound in \cref{sec:theory} to study the scaling of the sample complexity with respect to the dimension $d$ for the function in \cref{eqn:fx}.  
\Rev{The two-layer network in \cref{eqn:two_layer} can be viewed as a particular realization of GN.} According to \cref{thm:apost} and \cref{thm:apriori}, we need to estimate the Barron norm $\norm{f^*}_{\mc{B}}$. 

For a given function, the Barron norm is often difficult to compute due to the minimization with respect to all possible $(a,\pi)$. Instead we may compute the spectral norm. For a given function
\(f \in C(\Omega),\) let \(F\) be an extension of \(f\) to
\(\mathbb{R}^{d},\) denoted by $F\vert_\Omega=f$. Define the Fourier transform \[
\hat{F}(\vk)=\frac{1}{2\pi}\int_{\RR^d}e^{-i\vk\cdot\vx}F(\vx)\ud \vx.
\]
Then spectral norm of \(f\) is defined as
\begin{equation}
\norm{f}_{s}=\inf_{F\vert_\Omega=f} \int_{\mathbb{R}^{d}}\|\vk\|_{1}^{2}|\hat{F}(\vk)| \ud \vk.
\label{eqn:spectral_norm}
\end{equation}
Note that the infimum is taken over all possible extensions \(F\).
Then the Barron norm can be bounded by the spectral norm as \cite[Theorem 2]{EMaWu2019}
\begin{equation}
\|f\|_{\mathcal{B}} \leq 2 \norm{f}_s+2\|\nabla f(0)\|_{1}+2|f(0)|.
\label{eqn:barron_spectral}
\end{equation}
Therefore we may obtain an upper bound of the Barron norm via the spectral norm.

Let us now consider $f(\vx)$ in \eqref{eqn:fx_scaled}, which satisfies $\|\nabla f(0)\|_{1}=|f(0)|=0$. To evaluate the spectral norm, we consider the one-dimensional version \(g(x)=x^2\). Consider any \(C^2\) extension of \(g\) to \(\RR\), denoted by \(G\), which satisfies \(\int_{\RR} k^2 \hat{G}(k)\ud k<\infty\). Then by definition
\(\gamma_s(g) \le \int_{\RR} k^2 \hat{G}(k)\ud k<\infty\).

Now consider the extension \(F(\vx)=\frac{1}{d}\sum_{i=1}^d G(x_i)\), and
\(\hat{F}(\vk)=\frac{1}{d}\sum_{i=1}^d \hat{G}(k_i)\prod_{j\ne i}\delta(k_j)\). Here $\delta(\cdot)$ is the Dirac-$\delta$ function.
Then
\[
\int_{\RR^{d}}\|\vk\|_{1}^{2}|\hat{F}(\vk)| \ud \vk=\frac{1}{d}\sum_{i=1}^d \int_{\RR} k^2 \hat{G}(k)\ud k=\int_{\RR} k^2 \hat{G}(k)\ud k,
\]
which gives
\begin{equation}
\norm{f}_s\le  \int_{\RR} k^2 \hat{G}(k)\ud k.
\label{eqn:}
\end{equation}
Combined with \cref{eqn:barron_spectral}, we find that $\norm{f}_\mc{B}\sim\Or(1)$. Thus according to \cref{prop:bound_path} we expect that the path norm of the regularized solution satisfies $\|\hat{\theta}\|_{\mathcal{P}}\sim \Or(1)$.  Hence the leading term of the generalization gap is 
\begin{equation}
\|\theta\|_{\mathcal{P}}\sqrt{\frac{\ln (d)}{n}}\sim \sqrt{\frac{\ln (d)}{n}}.
\label{eqn:rate_generalization_error}
\end{equation}

This scaling seems quite favorable, as $n$ only needs to grow as $\ln(d)$. However, recall the discussion in \cref{sec:intro} that the target generalization error should be $\epsilon/d^2$, this means that the required sample complexity (up to logarithmic factors) is 
\begin{equation}
n\sim \Or(d^4/\epsilon^2).
\label{eqn:sample_complexity}
\end{equation}
According to \cref{thm:apriori}, the network also needs to be wide enough as $m\sim \Or(d^2/\epsilon)$. 
\(\lambda\sim \Or(n^{-\frac12})=\Or(\epsilon/d^2)\).
In particular, the sample complexity increases very rapidly with respect to the dimension $d$ in order to approximate the unscaled function $\wt{f}^*(\vx)$. 

The analysis of the local network (LN) is essentially applying a two-layer network $g(x;\theta)$\ to the scalar mapping $g(x)=x^2$. \Rev{Since $g(x)=x^2$ is a special case of $f^*(\vx)=\frac{1}{d}\sum_{i=1}^d x_i^2$ with $d=1$, by Eq.~\eqref{eqn:sample_complexity} and setting $d=1$, the sample complexity for the scalar mapping should be $n\sim \Or(1/\epsilon^2)$}. 
Let \(z_i=x_i^2-g(x_i;\theta)\),
and assume that the error from each component are of mean zero and  independent, i.e. 
\begin{equation}
\mathbb{E}(z_iz_j)=0, \quad i\ne j.
\label{eqn:error_condition}
\end{equation}
Then the generalization error is simply
\begin{align*}
L(\hat{\theta})=&\frac{1}{2}\mathbb{E}\left(\frac{1}{d}\sum_{i=1}^d x_i^2-\frac{1}{d}\sum_{i=1}^d g(x_i;\theta)\right)^2, \\
=&\frac{1}{2d^{2}} \mathbb{E}\left(\sum_{i=1}^d z_i^2+2\sum_{i\ne j}z_i z_j\right),\\
\approx& \frac{1}{2d} \mathbb{E}\left(z_i^2 \right) \sim \frac{1}{d}.
\end{align*}
Therefore the generalization error of LN can be $\Or(d^{-1})$ smaller than that of GN. Note that the condition \eqref{eqn:error_condition} is crucial for the $d^{-1}$ factor. In fact if the errors from all components are correlated, there may be no gain at all in terms of the asymptotic scaling with respect to $d$! 

However, our numerical results in \cref{sec:numer} demonstrate that the performance of LN can be significant better than GN by a very large margin. Therefore there is still gap in terms of the theoretical understanding of the performance of LN. 

\section{Numerical results}\label{sec:numer}

In this section we describe in detail the numerical experiments along with the analysis of the empirical results.

Our first goal is to study the dependence of the generalization or test error with respect to the architectural bias. In section~\ref{sec:architecture}, we showcase the superior performance of the local network, which is built using structural information about the underlying problem, compared to the  the global network, in which no information is used. 

Our second goal is to study the impact of the explicit regularization (versus implicit regularization) when training the global network to approximate the unscaled target function. In section~\ref{sec:sample_complexity}, we characterize the scaling of the test loss with respect to the input dimension, and the scaling of the test loss with respect to the number of samples with/without the explicit regularization. The explicit regularization improves both rates according to the numerical experiments.

\subsection{Data generation and loss function}
For all the numerical experiments, the dimension of the input denoted by $d$, ranges from $4$ to $60$ with step size $4$. For each $d$, we generate $10^6$ samples by first sampling $10^6 \times d$ numbers uniformly from the interval $[-1,1]$ and then organizing the resulting data as a matrix of dimensions $10^6 \times d$. We then compute $y = \sum_{i=1}^d x_i^2$ for each row. With this setup, the domain of the target function is restricted to $\Omega = [-1,1]^d$. 
 
We also generate data for the sum of quartic and the cosine terms, namely $y = \sum_{i=1}^d x_i^4$ or $y = \sum_{i=1}^d \cos{x_i}$. These datasets are used in section~\ref{sec:architecture} to showcase the importance of the architectural bias for target functions other than $y = \sum_{i=1}^d x_i^2$.

For the sake of reproducibility, all the experiments described below  use the data generated with a fixed seed.  In this section we use  $n$ to denote the number of the training samples and $N_{\text{sample}}$ for the number of total samples (training, validation, test all combined). For experiments that involve $N_{\text{sample}} \leq 10^6$ samples, we extract the first $N_{\text{sample}}$ rows from the total $10^6$ rows of data.

To simplify the training procedure, we enforce the permutation symmetry of the inputs by including the sorting procedure while preprocessing the input data. We point out that this permutation symmetry can be directly embedded in the network \cite{ZaheerKotturRavanbakhshEtAl2017}.

On one hand, we have done our analysis in section \ref{sec:theory} and \ref{sec:LN_GN} with respect to the scaled target function $f^*(\vx)$ in Eq.~\eqref{eqn:fx_scaled}, so the neural networks we build in this section aim to learn the scaled target function. Thus, the mean squared error loss for training data when optimizing the networks are defined as
\begin{equation}
    \text{MSE} = \frac{1}{n}\sum_{i=1}^{n} \left(f_{NN}(\vx^{(i)}, \theta) - f^*(\vx^{(i)})\right)^2
\label{eqn:MSE}
\end{equation}
Here $n$ denotes the number of training data, $f_{NN}$ denotes the function represented by the network, and $\vx^{(i)}$ represents $i$th row in the dataset. The test loss can be calculated in the same way using the test data instead of the training data.
On the other hand, we are interested in the performance of the models in the original scale, so the reported training/test loss are scaled by $d^2$ to represent the mean squared error of approximating the original unscaled target function:
\begin{equation}
\text{MSE}_{\text{originial}}=\frac{1}{n}\sum_{i=1}^{n} \left(d\cdot f_{NN}(\vx^{(i)}, \theta) - \wt{f}^*(\vx^{(i)})\right)^2
\label{eqn:MSE_unscaled}
\end{equation}

\subsection{Architectural bias}
\label{sec:architecture}

In this section we showcase the empirical effects of the architectural bias on the test or generalization errors. To illustrate this effect, in a slightly more generality, we consider as target functions: $\wt{f}^*(\vx):=\sum_{i=1}^d x_i^2$, $\sum_{i=1}^d x_i^4$, and $\sum_{i=1}^d \cos x_i$, respectively. We use three different network architectures to approximate each target function: global network, the local network, and the locally connected network (LCN), which is similar to the local network, but whose weights are not longer shared.

\begin{figure}[ht] 
\centering
\includegraphics[width=0.95\textwidth]{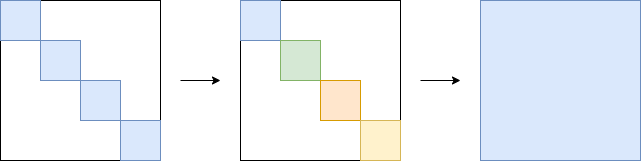}
\caption{The relation of the weight matrices after embedding weight matrices for all three networks (between the first hidden layer and the second hidden layer with input dimension $d=4$) into the weight matrix for the global network. The arrows mean that the matrices on the left hand side are subsets of the matrices on the right hand side. The blocks in the first weight matrix have the same color to illustrate the weight-sharing. Matrix elements in the white regions are zeros.} 
\label{fig:embedding} 
\end{figure}

These three architectures are closely related. In the global (or dense) network, each layer is represented by a weight matrix that is an arbitrary dense matrix (see Figs.~\ref{fig:architecture} and \ref{fig:embedding}). In the locally connected network we have the same matrix but constrained to be block diagonal. Finally, in the local network we further constraint the weight matrix to be both block diagonal and block circulant, thus implying that the block in the diagonal are the same (usually called weight-sharing in the machine learning community). This observation allows us to embed a local network into a locally connected one, and then into a global one. This embedding is performed by simply copying the corresponding entries of the weights matrices at each layer and filling the rest with zeros as shown in Fig.~\ref{fig:embedding}. 

Following \cite{Hornik91} all three networks can approximate the target function to arbitrary accuracy with just one hidden layer: the result is straightforward for the global network by the universal approximation; for the locally connected and local networks, the result stems from applying the universal approximation to each coordinate, thus approximating the scalar component function $g: \mathbb{R} \to \mathbb{R}$ ($g(x)=x^2$ in the square case, $x^4$ in the quartic case and $\cos x$ in the cosine case). In practice, however, we find that with two hidden layers the networks are easier to train, all the hyperparameters being equal.

\textbf{}
Following these considerations, for all the numerical experiments we fix the number of hidden layers to be two. In addition, we define $\alpha$ as the number of nodes per input node, which for the local networks coincides with the number of channels. Thus we have $d\cdot \alpha$ nodes in the hidden layers for the global network.

The networks are implemented\footnote{Detailed implementation can be found at \url{https://github.com/jfetsmas/NormSquareLearning.git}} with Keras \cite{keras}, using Tensorflow \cite{tensorflow} as the back-end. Within the Keras framework, we use dense layers, locally connected layers and one-dimensional convolutional layers to implement the global network, the locally connected network, and the local network, respectively (see Algs.~\ref{alg:globalConNet}, \ref{alg:locallyConNet}, and \ref{alg:localNet}). We add a lambda layer that divides the output by $d$ at the end of the networks to learn the scaled function $f^*(\vx)$ in \cref{eqn:fx_scaled}.
For both the locally connected and convolutional layers, we use stride and window size both equal to one, and $\alpha$ the number of channels. We point out that $\alpha$ can be considered as the number of nodes in the hidden layers for the block component that approximates the component function $g$  as shown in Fig.~\ref{fig:architecture}. In this way, the function represented by the local network has the structure $\frac{1}{d}\sum_{i=1}^d g(x_i, \theta)$ enforced by the architecture, where $g(\cdot, \theta)$ is the neural network block that approximates component function $g$.

\begin{algorithm}[H]
\SetAlgoLined
\begin{verbatim}
layerInput       = Input(shape=(d,))
layerHidden1     = Dense(DenseNodes, activation='relu',
                         use_bias=True)(layerInput)
layerHidden2     = Dense(DenseNodes, activation='relu',      
                         use_bias=True)(layerHidden1)
layerOutput_pre  = Dense(1, activation='linear', use_bias=False)(layerHidden2)
layerOutput      = Lambda(lambda x: x / d)(layerOutput_pre)
model            = Model(inputs=layerInput, outputs=layerOutput) 
\end{verbatim}\caption{Code for the global network} \label{alg:globalConNet}
\end{algorithm}

\begin{algorithm}[H]
\SetAlgoLined
\begin{verbatim}
layerInput   = Input(shape=(d,1))
layerHidden1 = LocallyConnected1D(alpha, 1, strides=1, 
               activation='relu', use_bias=True)(layerInput)
layerHidden2 = LocallyConnected1D(alpha, 1, strides=1, 
               activation='relu', use_bias=True)(layerHidden1)
layerOutput  = LocallyConnected1D(1, 1, strides=1, 
               activation='linear', use_bias=False)(layerHidden2)
Sum          = Lambda(lambda x: K.sum(x, axis=1), name='sum')
layerSum_pre = Sum(layerOutput)
layerSum     = Lambda(lambda x: x / d)(layerSum_pre)
model        = Model(inputs=layerInput, outputs=layerSum)  
\end{verbatim}\caption{Code for the locally connected network} \label{alg:locallyConNet}
\end{algorithm}

\begin{algorithm}[H]
\SetAlgoLined
\begin{verbatim}
layerInput   = Input(shape=(d,1))
layerHidden1 = Conv1D(alpha, 1, strides=1, activation='relu', 
               use_bias=True)(layerInput)
layerHidden2 = Conv1D(alpha, 1, strides=1, activation='relu', 
               use_bias=True)(layerHidden1)
layerOutput  = Conv1D(1, 1, strides=1, activation='linear', 
               use_bias=False)(layerHidden2)
Sum          = Lambda(lambda x: K.sum(x, axis=1), name='sum')
layerSum_pre = Sum(layerOutput)
layerSum     = Lambda(lambda x: x / d)(layerSum_pre)
model        = Model(inputs=layerInput, outputs=layerSum) 
\end{verbatim} \caption{Code for the local network}\label{alg:localNet}
\end{algorithm}

In the experiments, we set the number of channels $\alpha = 50$ and we use ReLU as the nonlinear activation function. The models are permutation invariant due to a sorting procedure used to pre-process the data. In all experiments we use the default weight initializer (glorot uniform initializer) and the Adam optimizer with starting learning rate = 0.01 and other default parameters \cite{adam}. 

We split the data into training, validation, and test datasets. Among the data for a single numerical experiment, $64\%$ is training data, $16\%$ is  validation data, and $20\%$ is test data. We use batch size = (number of training and validation samples) / 100 for all experiments in this subsection, and thus each epoch contains 80 iterations.
During the training process, we evaluate the loss on the validation set every epoch and we keep the model with the lowest validation loss, which is then reported.

\begin{figure}[ht] 
\centering
{
\subfigure[target function $y=\sum_{i=1}^d x_i^2$]{%
\label{fig:Compare_square}
\includegraphics[width=0.3\textwidth]{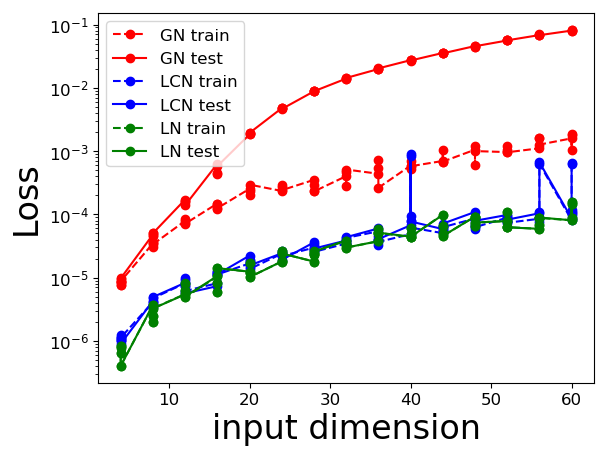}
}
\subfigure[target function $y=\sum_{i=1}^d x_i^4$]{%
\label{fig:Compare_quartic}
\includegraphics[width=0.3\textwidth]{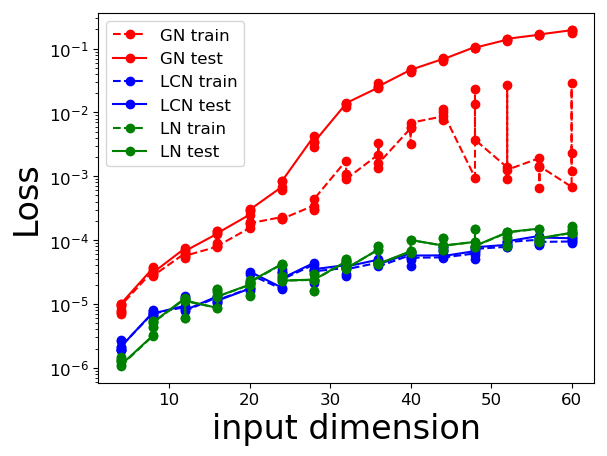}
}
\subfigure[target function $y=\sum_{i=1}^d \cos x_i$]{%
\label{fig:Compare_cosine}
\includegraphics[width=0.3\textwidth]{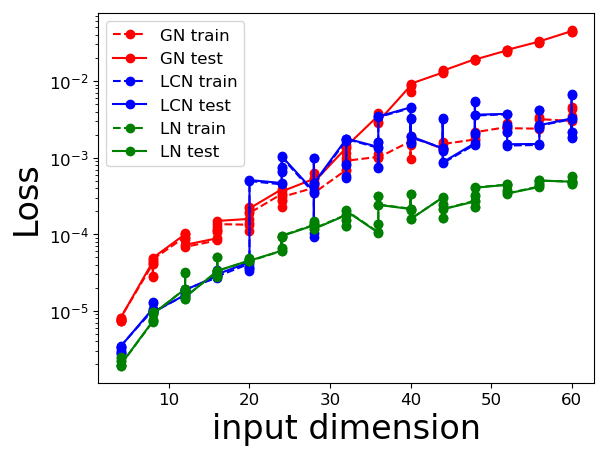}
}
\caption{Comparison of rescaled average mean square error for the three architectures. GN stands for global network, LCN for locally connected network, and LN for local network.}
\label{fig:Compare3arch} 
}
\end{figure}

For each value of the input dimension, we run four experiments with the same configuration, but with a different random seed for the optimizer. All experiments in this subsection use a dataset of size $N_{\text{sample}}=10^5$ (training, validation, and test data all combined).

Fig.~\ref{fig:Compare3arch} depicts the behavior of the test and training losses for the three networks as the input dimension increases. The losses are computed using the mean squared error in the original scale as in Eq.~\eqref{eqn:MSE_unscaled}.
For all three target functions, the local network significantly outperforms the global network, especially for large input dimension. LN and LCN shows comparable performance for the quadratic and quartic functions. The training error of LCN is larger for the cosine function, but there is no noticeable generalization gap. We expect that the performance of LCN can be further improved through further hyperparameter tuning. In all three cases, the global network exhibits a large generalization gap, whereas this gap is almost non-existent for the local network. These results clearly indicates the influence of the architecture in the accuracy of the approximation and the generalization gap.

\begin{figure}[ht] 
\centering
\includegraphics[width=0.5\textwidth]{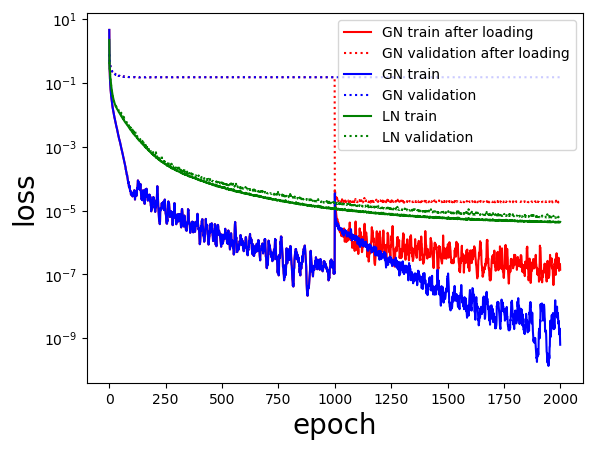}
\caption{Evolution of the mean squared error  with respect to the number of epochs for the different networks. Blue lines are the trajectories of the training and validation loss of the global network; green lines are for the local network; Red lines are the trajectories after loading the local network weights into the global network at epoch 1000.}
\label{load} 
\end{figure}   
As depicted in Fig.~\ref{fig:embedding}, we know that the local networks are indeed a subset of the global ones, and given the high-level of accuracy obtained by the local network we can infer that the global network should have enough representation power to approximate the target solution accurately. Following this logic, the large gap can be then attributed to the optimization procedure, which is unable to identify a suitable approximation to the local network (``needle'') among all overparameterized dense networks (``haystack''). 

A natural question is: can the optimization procedure find a better solution starting from a good initial guess? To study this, we train a local network until a local minimum is achieved, and we transfer this minimum to its corresponding global network, and we use it as an initial guess for the training procedure. In particular, we fix $d= 20$, and the number of data at $1000$. We train both the local and the global networks for $1000$ epochs, do the weight transfer, and then resume training for another $1000$ epochs. To obtain more stable training curves, we add decay of $0.03$ in the argument of the Adam optimizer. 
From Fig.~\ref{load} we can observe that by properly initializing the global network, we are able to significantly reduce the test loss, and to drastically bridge the generalization gap (red lines starting at epoch 1000). Although the training error of the global network is consistently lower than that of the local network  (solid blue and red lines, compared with solid green line), the generalization gap (the gap between solid and dashed for the blue and red lines, compared with the gap for the green lines) of the global network remains noticeably larger compared with the local network.

\subsection{The impact of the explicit regularization}
\label{sec:sample_complexity}

In this subsection, we explore the relation between the test loss, the number of samples and the input dimension. The calculations in section \ref{sec:LN_GN} suggests that the leading term of the generalization gap should satisfy Eq.~\eqref{eqn:rate_generalization_error}. For the global network, the generalization gap is sufficiently large\footnote{Given that the training loss is usually one or more magnitude smaller than the test loss.} that we can regard the test loss to be approximately equal to the generalization error. Thus, we expect the test loss of the global network to be bounded by Eq.~\eqref{eqn:rate_generalization_error} (multiplied by $d^2$ for the unscaled function). We test the tightness of the bound by conducting the following two experiments: 
\begin{enumerate}
    \item we fix the sample size and investigate the relation between the test loss and the input dimension; and
    \item we fix the input dimension $d$ and obtain the rate of growth of the test loss with respect to the sample size.
\end{enumerate}
\Rev{To summarize our numerical observations, for the first experiment}, despite numerical errors and the fact that we only use the leading term in Theorem~\ref{thm:apost}, we observe that the rate with respect to the input dimension suggested by Eq.~\eqref{eqn:rate_generalization_error} regarding the input dimension is close to optimal when using the explicit regularization. \Rev{For the second experiment however,} the rates with respect to the number of samples obtained in numerical experiments are around $\mathcal{O} (N_{\text{sample}}^{-0.8})$ without explicit regularization, and $\mathcal{O} (N_{\text{sample}}^{-1.0})$ with explicit regularization. Hence the convergence rate with respect to the number of samples is faster than the theoretical predicted worst case rate as $\mathcal{O} (N_{\text{sample}}^{-0.5})$.

\begin{figure}[htbp]
\centering
{
\subfigure[the relationship between the generalization error and the input dimension. Slope of the best fitted line (red) is 3.638]{%
\label{fig:testWoreg}
\includegraphics[width=0.45\textwidth]{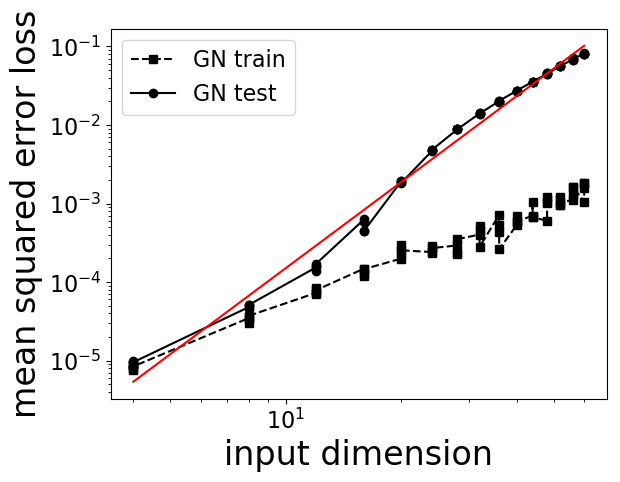}
}
\subfigure[the relationship between the path norm and the input dimension. Slope of the best fitted line (red) is 1.107]{%
\label{fig:pathnormWoreg}
\includegraphics[width=0.45\textwidth]{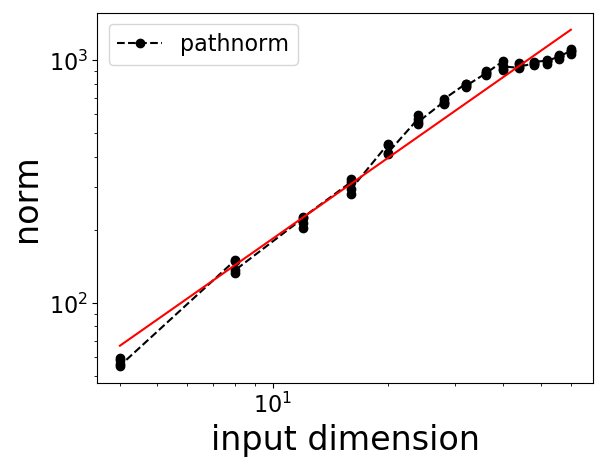}
}
\caption{Experiments without regularization}
\label{fig:Woreg} 
}
\end{figure}

\Rev{\textbf{Loss Vs. input dimension without explicit regularization}}\\
Here we fix the sample size to $10^5$. For each $d$ ranging from $4$ to $60$, we repeat the training procedure four times with the same set of hyperparameters (the default values have been discussed in section~\ref{sec:architecture}).
In Fig.~\ref{fig:testWoreg}, we display the training loss and test loss with respect to the input dimension in log-log scale. The losses are computed using the mean squared error in the original scale as in Eq.~\eqref{eqn:MSE_unscaled}.
The slope in Fig.~\ref{fig:testWoreg} indicates that the generalization error of the approximation to the original target function $\wt{f}^*(\vx)$ as in Eq.~\eqref{eqn:fx} grows as $d^{3.6}$, and thus the generalization error of the approximation to the scaled $f^*(\vx)$ as in Eq.~\eqref{eqn:fx_scaled} grows as $d^{1.6}$. The empirical rate $d^{1.6}$ is a lot larger than logarithmic growth predicted in the bound in Eq.~\eqref{eqn:rate_generalization_error}. 

In particular, Eq.~\eqref{eqn:rate_generalization_error} indicates that in order to bound the generalization error, the path norm needs to be $\sim \Or(1)$. However, the boundedness of the path norm is only proven in the context of explicit regularization. Without such an explicit regularization  term (also called the ``implicit regularization''), there is no \textit{a priori} guarantee that the path norm can remain bounded as the input dimension increases. 

We remark that to compute the path norm for our three-layer network, we need to modify the formula for two-layer network. Let us denote the weight matrices of the three layers to be $\vw^1, \vw^2, \vw^3$. With the width of hidden layer for the global network being $d\cdot \alpha$ as shown in section \ref{sec:architecture}, the size of the weight matrices are $d\cdot \alpha \times d, d\cdot\alpha \times d\cdot \alpha, 1 \times d\cdot \alpha$ respectively. In the three-layer case, the path norm can be calculated using the formula:
\begin{equation}
\|\theta\|_{\mathcal{P}} := \frac{1}{d}\sum_{i=1}^d\sum_{j,k=1}^{d\alpha}  |\vw^1_{ij}||\vw^2_{jk}||\vw^3_{k}|
\label{eqn:pathnorm_calc}
\end{equation}
The $\frac{1}{d}$ factor is due to the last layer in Alg.~\ref{alg:globalConNet}, which divides the output by $d$. Using the updated formula we plot the path norm with respect to the input dimension in Fig.~\ref{fig:pathnormWoreg}. From the figure we can observe that the path norm grows as $\sim d^{1.1}$, thus violating the $\Or(1)$ assumption for Eq.~\eqref{eqn:rate_generalization_error}.
Notice that in this scenario, although Eq.~\eqref{eqn:rate_generalization_error} does not apply, the bound in Theorem~\ref{thm:apost} provides a fairly good estimate of the growth of the generalization error. The leading term in the bound given by Theorem~\ref{thm:apost} grows as $\|\theta\|_{\mathcal{P}}\sqrt{\ln(d)}$. With the path norm $\sim d^{1.1}$, and $\sqrt{\ln(d)}$ empirically behaving like a fractional power of $d$ for small $d$, the product has a rate similar to the rate of the test loss observed, which is $\sim d^{3.6}$ (after taking into account the $d^2$ scaling factor). 

\Rev{\textbf{Loss Vs. input dimension with explicit regularization}}\\
To demonstrate that explicit regularization indeed reduces the path norm, test loss and generalization error, we implement three types of regularization schemes, and study the growth of the errors with respect to the input dimension.
Let the number of trainable parameters in the neural network be $N_{\text{par}}$. Denote the trainable parameters by $\{\theta_i\}_{i=1}^{N_{\text{par}}}$ and regularization constant by $\lambda$. We can summarize the implementation of the three regularization as
\begin{itemize}
    \item L1 regularization, minimizing MSE+$\lambda\sum_{i=1}^{N_{\text{par}}} |\theta_i|$
    \item L2 regularization, minimizing MSE+$\lambda\sum_{i=1}^{N_{\text{par}}} \theta_i^2$ 
    \item path norm regularization, minimizing MSE+$\lambda \|\theta\|_{\mathcal{P}}$, where $\|\theta\|_{\mathcal{P}}$ is calculated by Eq.~\eqref{eqn:pathnorm_calc}
\end{itemize}

\begin{figure}[htbp]
\centering
{%
\subfigure[Loss Vs. input dimension for GN with L1 regularization. Slope of the best fitted line (yellow): 2.527]{%
\label{fig:regl1}
\includegraphics[width=0.3\textwidth]{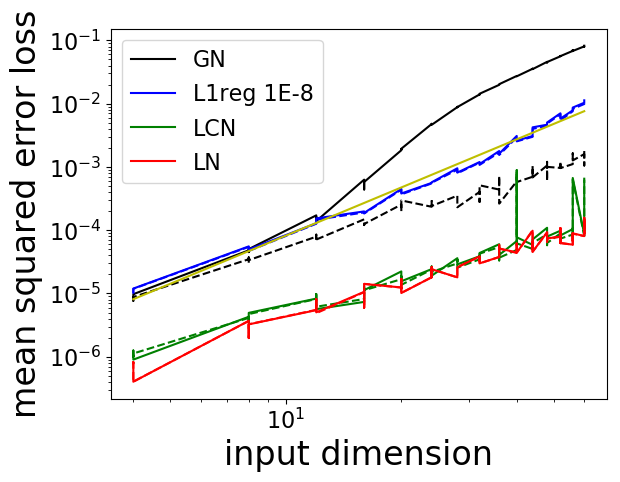}
} 
\subfigure[Loss Vs. input dimension for GN with L2 regularization. Slope of the best fitted line (yellow): 3.051]{%
\label{fig:regl2}
\includegraphics[width=0.3\textwidth]{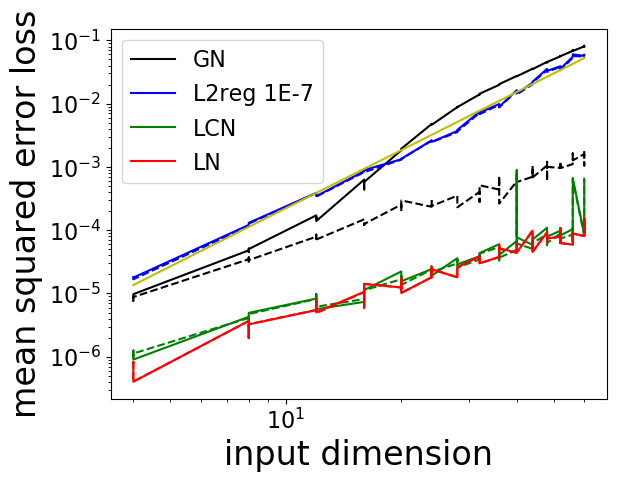}
}
\subfigure[Loss Vs. input dimension for GN with path norm regularization. Slope of the best fitted line (yellow): 2.098]{%
\label{fig:regpath}
\includegraphics[width=0.3\textwidth]{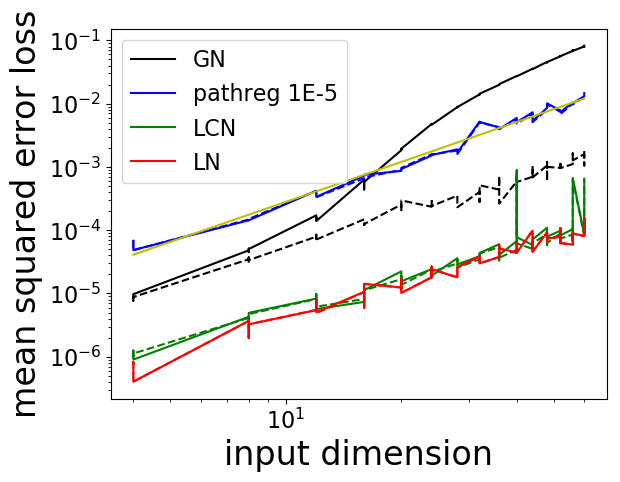}
}
\caption{Experiments with regularization, solid lines are test loss and dashed lines are training loss.}
\label{fig:reg}
}
\end{figure}

The MSE is computed as in Eq.~\eqref{eqn:MSE}.
To implement the three regularization schemes, we use the kernel regularizer in Keras in tensorflow 1.7.0 for the L1 and the L2 regularization (the regularization is only on the weight matrices, but not on the bias), and we use tensorflow 2.0 to implement the path norm regularization. We choose the regularization constant $\lambda$ (fixed with respect to $d$) empirically so that the global network achieves the best test loss. The regularization constants we select in Fig.~\ref{fig:reg} are $10^{-8}$ for L1 regularization, $10^{-7}$ for the L2 regularization, and $10^{-5}$ for the path norm regularization.

In addition to the test/train loss Vs. the input dimension for the regularization experiments, we also display our previous results for GN, LN, and LCN as a reference in Fig.~\ref{fig:reg}

\begin{figure}[ht] 
\centering
\includegraphics[width=0.5\textwidth]{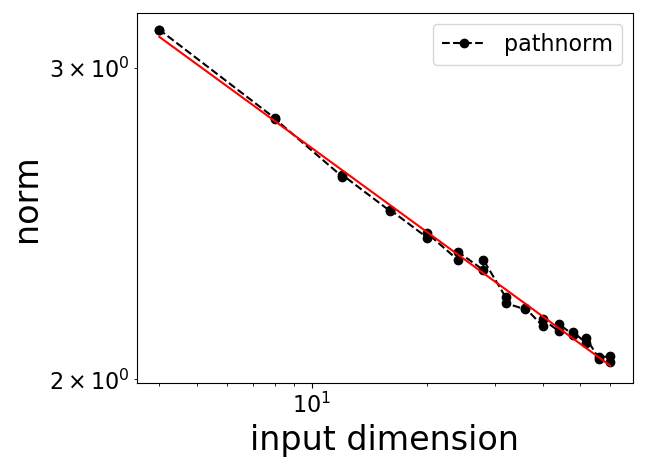}
\caption{the relationship between the path norm and the input dimension for the path norm regularization experiment with $\lambda = 10^{-5}$. Slope of the best fitted line (red): -0.158}
\label{fig:pathnorm_parhreg} 
\end{figure}

Recall that the growth rate of the test loss for the global network without regularization (shown in Fig.~\ref{fig:testWoreg}) is around 3.638, all three regularization helps reduce down the rate in Fig.~\ref{fig:reg}. Path norm regularization in Fig.~\ref{fig:regpath} exhibits the best growth rate, despite that the test loss for small $d$ is slightly larger.  We plot the path norm Vs. the input dimension in Fig.~\ref{fig:pathnorm_parhreg} and we indeed observe that the path norm is $\Or(1)$ as desired. As a matter of fact, the path norm even decreases slightly as $d$ increases, and this is qualitatively different from the behavior of implicit regularization. In addition, with the explicit regularization and thus bounded path norm, Eq.~\eqref{eqn:rate_generalization_error} gives a tight estimate of the rate of growth.

Since the weight matrices for the LN embedded GN are block diagonal, and hence are sparse matrices. We will look at the first weight matrices of the models under the following scenario:
\begin{itemize}
    \item GN without regularization,
    \item GN with L1 regularization,
    \item GN with L2 regularization,
    \item GN with path norm regularization, and
    \item GN with optimal LN weights embedded.
\end{itemize}

\begin{figure}[htbp]
\centering
{%
\subfigure[GN w/o reg]{%
\label{fig:l1densematrix}
\includegraphics[height=0.28\textheight]{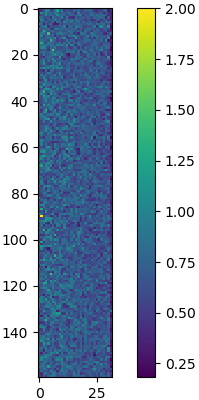}
} 
\subfigure[GN L1 reg]{%
\label{fig:l1sparsematrix}
\includegraphics[height=0.28\textheight]{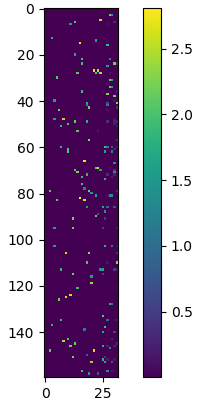}
}
\subfigure[GN L2 reg]{%
\label{fig:l1densematrix}
\includegraphics[height=0.28\textheight]{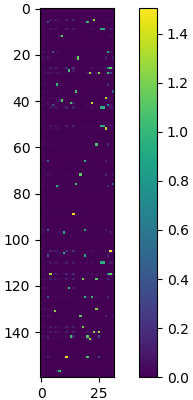}
} 
\subfigure[GN path norm reg]{%
\label{fig:l1sparsematrix}
\includegraphics[height=0.28\textheight]{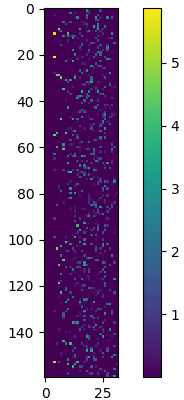}
}
\subfigure[loaded LN]{%
\label{fig:l1densematrix}
\includegraphics[height=0.28\textheight]{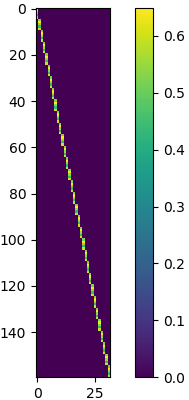}
} 
\caption{The sparsity pattern of the weight matrices between the first and second layers after training.}
\label{fig:sparseweight}
}
\end{figure}

In Fig.~\ref{fig:sparseweight}, we display the first weight matrices for input dimension $d=32$ for the five trained models. Since the weight matrix is from the input nodes to the first hidden layer, the dimension is $1600\times 32$, where 1600 is obtained from input dimension (32) multiplied by number of nodes per input node (50). To improve visibility, we display the maximum absolute value among the 10 adjacent cells so that the first dimension is reduced by a factor of 10. We can see from the picture that L1 regularization, L2 regularization, and path norm regularization all lead to a much sparser weight matrix compared with the global network without regularization.

\begin{figure}[htbp] 
\centering
{
\subfigure[Loss Vs. sample size for GN with fixed batch size. Slope of the best fitted line for test loss with an obvious generalization gap (red): -0.745]{%
\label{fig:EvsSample_fixedbatch}
\includegraphics[width=0.45\textwidth]{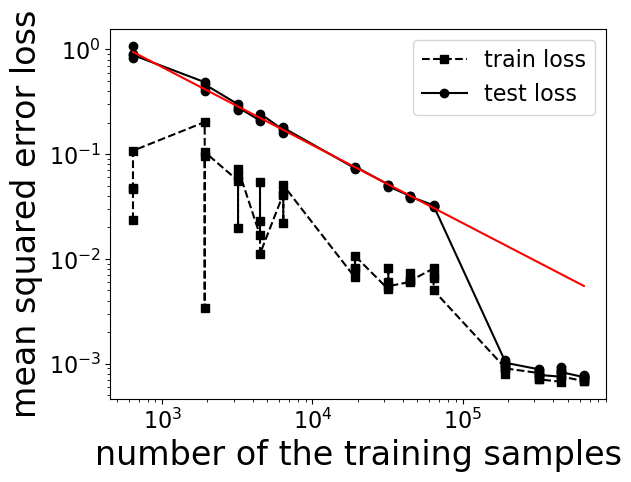}
} 
\subfigure[Loss Vs. sample size for GN with batch size given by a fixed ratio. Slope of the best fitted line (red) is -0.820.]{%
\label{fig:EvsSample_fixedratio}
\includegraphics[width=0.45\textwidth]{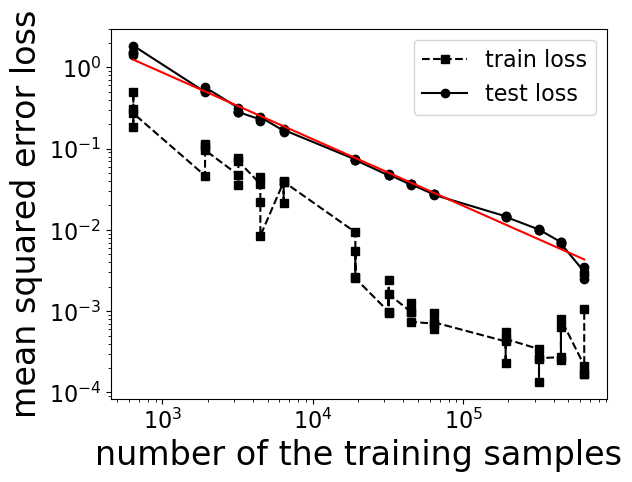}
}
\caption{the relationship between the test loss and the sample size with two choices of batch size.}

\label{EvsSample_batch} 
}
\end{figure}

\Rev{\textbf{Loss Vs. sample size without explicit regularization}}\\
For the experiment regarding the rate of growth of the test loss with respect to the sample size, we fix $d = 40$ (the choice is arbitrary), and we vary the sample size from $10^3$ to $10^6$. Since the sample size varies, we consider the following two choices of the batch size: 
\begin{enumerate}
\item fix the ratio and let batch size be (number of training and validation samples) / 100, this choice ensures same number of iterations per epoch as the sample size increases.
\item fix the batch size to 80 as sample size varies. The number of iterations increases as the sample size increases.
\end{enumerate}  
The other hyperparameters are the same as the setup in section~\ref{sec:architecture}. We run 4 training procedures for each number of samples and we report the resulting test loss (early stopping still applies) versus the number of samples, which are summarized in Fig.~\ref{EvsSample_batch}. In Fig.~\ref{fig:EvsSample_fixedbatch}, with batch size fixed at 80, the performance changes when the number of training samples exceed $10^5$. Since we focus on the generalization error in this paper (so we need the existence of generalization gap), and also the time cost of a training procedure drastically increases when the small batch size is applied to a large sample size (number of iterations per epoch, computed by training sample size / batch size, is large), we focus on the segment where the number of training samples is below $10^5$ and the generalization gap is visible. Following Theorem~\ref{thm:apost}, the generalization error decreases asymptotically as $\mathcal{O} (N_{\text{sample}}^{-0.5})$, but numerically we observe different exponents in both figures, which is around $-0.8$. The fixed batch size experiment in Fig.~\ref{fig:EvsSample_fixedbatch} returns a similar rate to the fixed ratio experiment in Fig.~\ref{fig:EvsSample_fixedratio}, therefore we may keep using the default setup (batch size with fixed ratio) for the rest of the studies in this subsection.

\begin{figure}[htbp] 
\centering
\includegraphics[width=0.5\textwidth]{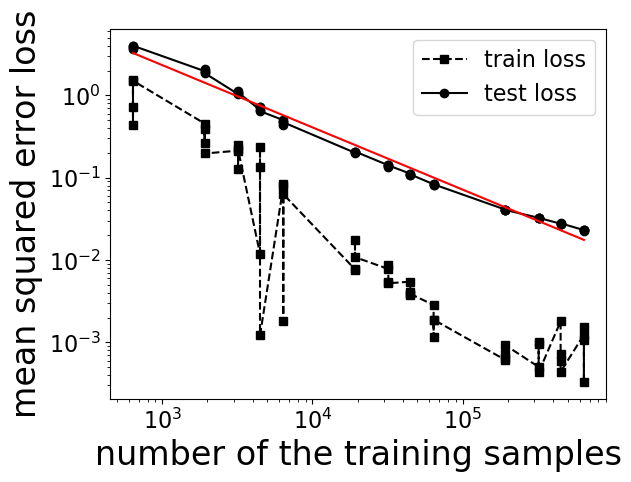}
\caption{Loss Vs. sample size for GN for $d=60$. Slope of the best fitted line (red) is -0.756} 
\label{fig:EvsSample_noreg} 
\end{figure}

To confirm that the rate is consistent for other input dimension values, we repeat the same procedure (with batch size obtained by the fixed ratio) for $d=60$. The trend and rate are reported in Fig.~\ref{fig:EvsSample_noreg}, and indeed the rate is still around $-0.8$, different from the theoretical rate $-0.5$.

\begin{figure}[htbp] 
\centering
{
\subfigure[Loss Vs. sample size for GN with L1 regularization for $d=40$. Slope of the best fitted line for test loss with an obvious generalization gap (red): -0.943]{%
\label{fig:EvsSample_l1reg_d40}
\includegraphics[width=0.45\textwidth]{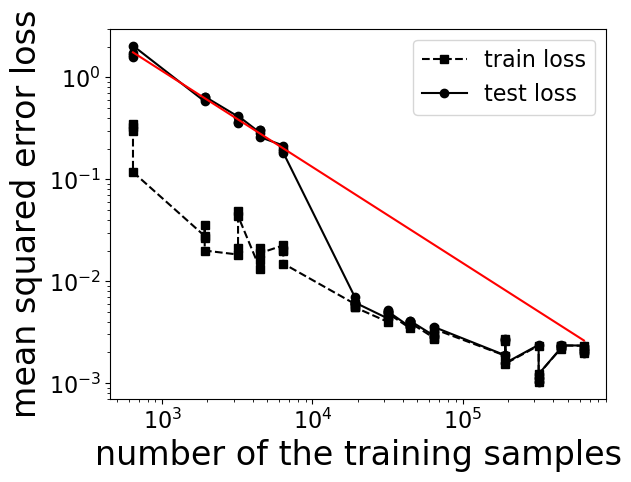}
} 
\subfigure[Loss Vs. sample size for GN with L1 regularization for $d=60$. Slope of the best fitted line for test loss with an obvious generalization gap (red): -0.986]{%
\label{fig:EvsSample_l1reg_d60}
\includegraphics[width=0.45\textwidth]{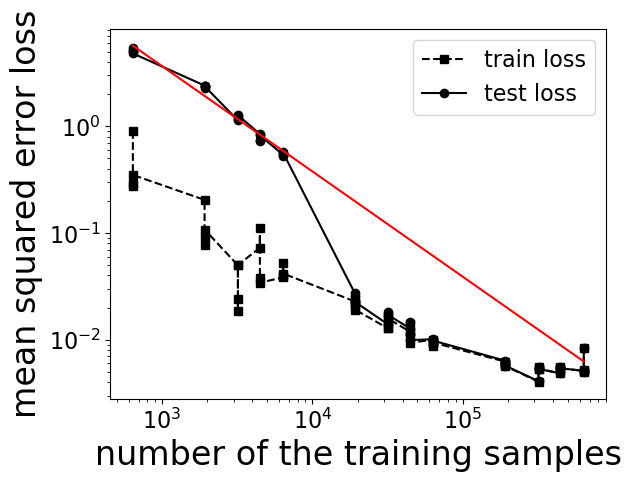}
}
\caption{the relationship between the test loss and the sample size for the global network with L1 regularization.}
\label{EvsSample_l1reg} 
}
\end{figure}

\Rev{\textbf{Loss Vs. sample size with explicit regularization}}\\
We test the impact of explicit regularization on the rate of growth of the test loss with respect to the sample size by adding L1 regularization to all the weight matrices. The setup is the same as the global network with L1 regularization experiment displayed in Fig.~\ref{fig:regl1}, except that instead of fixing $N_{\text{sample}}=10^5$ and letting $d$ varies, we fix $d=40$, $d=60$ and let $N_{\text{sample}}$ vary from $10^3$ to $10^6$. We fix the regularization constant at $10^{-8}$ as in Fig.~\ref{fig:regl1}. The rates of the segment where there is a visible generalization gap are shown in Fig.~\ref{EvsSample_l1reg}. We observe that as $N_{\text{sample}}$ increases to certain point (around $10^4$), the fixed regularization constant makes the regularization term more significant in the loss, resulting  in nearly vanishing generalization gap. Since our focus is on generalization error inspired by the theory in section~\ref{sec:theory} and \ref{sec:LN_GN}, we report the rate in the segment with visible generalization gap. Notice that the rate in the L1 regularization is around $-1.0$, and hence the decay of the test loss with respect to the number of samples is faster than that without explicit regularization, which is around $-0.8$. 

\Rev{\textbf{Empirical sample complexity with and without explicit regularization}}\\
Eq.~\eqref{eqn:sample_complexity} gives a theoretical prediction of the scaling of growth of number of samples with respect to the input dimension in order to maintain a predetermined level of test loss. We find this relation hard to verify directly because: 1) The input dimension, and the number of samples are usually chosen among finitely many values, so the grid may not be fine enough to find the $(n,d)$ combination that gives the predetermined test loss; 2) The test loss varies even with same hyperparamter setting, so it is rather difficult to make sure that the test loss stays at a constant value especially given that the loss of the scaled function can be very small ($10^{-6}\sim 10^{-4}$). However, we can compute the numerical sample complexity if we assume that the generalization gap is in the function form $\frac{d^{\beta_1}}{n^{\beta_2}}$ as in Eq.\eqref{eqn:rate_generalization_error}. Recall that theoretical prediction gives $\beta_1 = 2$ because the loss for the original function needs to be scaled by $d^2$ as noted in Eq.\eqref{eqn:MSE}, and $\beta_2 = 0.5$. Then the sample complexity with respect to $d$ can be characterized by a rate $\gamma$ as $n \sim \Or(d^{\gamma})$, where $\gamma = \frac{\beta_1}{\beta_2}=4$.

The rate we obtained for the implicit regularization is $\beta_1 = 3.6$ as in Fig.~\ref{fig:testWoreg}, and $\beta_2 = 0.8$ as in Fig.~\ref{fig:EvsSample_fixedratio}, and thus the sample complexity rate is approximately $\gamma = 4.5$, i.e. $n \sim \Or(d^{4.5})$. This is close to the theoretically predicted rate, but this largely benefited from the observation that $\beta_2=0.8>0.5$. The rate we obtained for the explicit regularization (L1 with regularization constant $10^{-8}$ to be precise) is $\beta_1 = 2.5$ as in Fig.~\ref{fig:regl1}, and $\beta_2 = 1.0$ as in Fig.~\ref{EvsSample_l1reg}, and thus the sample complexity rate is approximately $\gamma=2.5$, i.e. $n \sim \Or(d^{2.5})$, which is significantly better than the theoretically predicted rate. We do point out that this is a very rough estimate because of the assumed function form and the fact that the $\beta_1$ rate in the L1 regularization in Fig.~\ref{fig:regl1} is not based on the generalization gap but the test loss (the generalization gap is very small). Overall, the explicit regularization helps improve the rate in both the test loss Vs. the input dimension, and the test loss Vs. the number of training samples, so the advantage of the explicit regularization is convincing.

\section{Discussion}\label{sec:conclusion}

Despite the fact that an overparameterized neural network architecture can represent a large class of functions, such representation power can come at the cost of a large sample complexity. This is particularly relevant in many scientific machine learning applications, as the required accuracy (in the form of a regression problem) is high, and the training data can be difficult to obtain. Therefore a number of recent works have focused on domain-specific neural network architectures aiming at reducing the number of parameters, ``retreating'' from the overparameterized regime. 

This paper gives an unambiguous, and minimal working example illustrating why this makes sense. Even for the seemingly simple task of computing the sum of squares of $d$ numbers in a compact domain (i.e. learning the square of a 2-norm), a general purpose dense neural network struggles to find a highly accurate approximation of the function even for relatively low $d$ (tens to hundreds). In particular, the sample complexity of an empirically optimally tuned and explicitly regularized dense network is $\Or(d^{2.5})$. This behaves better than the sample complexity from \textit{a priori} error bound, which is  $\Or(d^4)$. The origin of such improvement deserves study in the future. The sample complexity  of an empirically optimally tuned and implicitly regularized dense network is close to $\Or(d^{4.5})$, and hence can be prohibitively expensive as $d$ becomes large. 

When we choose a proper architecture, such as the local network or the locally connected network, the generalization error still grows with respect to $d$. However, we find that the generalization gap is nearly invisible with explicit or implicit regularization, and the test loss is orders of magnitude smaller than that of the global (dense) network. Given that the sample complexity can asymptotically scale as the square of the test loss (assuming training error is negligible), the practical savings due to the use of a local network is vital to the success of the neural network. 

From a theoretical perspective, we remark that existing error analysis based on the Rademacher complexity-type cannot yet explain why the prefactor of the local network should be lower by orders of magnitude compared to the global (dense) network. Our results illustrate that implicit regularization as early stopping may not give the optimal generalization error rate in practice, although in theory it can be shown to achieve the same optimal rates as $L^2$ regularization in reproducing kernel Hilbert spaces \cite{onEarlyStopping,Wei2017boosting}. Given that implicit regularization is still a prevailing regularization method used in practical applications, better theoretical understanding of the behavior of early stopping regularization in neural networks, and methods to improve the performance of such an implicit regularization in practice are also needed.

\section*{Acknowledgments}

This work was partially supported by the Department of Energy under
Grant No. DE-SC0017867, the CAMERA program (L. L., J. Z., L. Z.-N.), and the Hong Kong Research Grant Council under Grant No. 16303817 (Y. Y.).
We thank the Berkeley Research Computing (BRC) program at the University
of California, Berkeley, and the Google Cloud Platform (GCP) for the
computational resources. We thank Weinan E, Chao Ma, Lei Wu for pointing
out the critical role of the path norm in understanding the numerical
behavior of the generalization error, and thank Joan Bruna, Jiequn Han,
Joonho Lee, Jianfeng Lu, Tengyu Ma, Lexing Ying for valuable
discussions.

\bibliographystyle{abbrv}
\bibliography{nn}

\end{document}